%% file: acl_latex.tex
\documentclass[11pt]{article}

\usepackage[final]{acl}

\usepackage{times}
\usepackage{latexsym}

\usepackage[T1]{fontenc}

\usepackage[utf8]{inputenc}

\usepackage{microtype}

\usepackage{inconsolata}

\usepackage{graphicx}
\usepackage{booktabs}
\usepackage{multirow}
\usepackage{subcaption}
\usepackage{siunitx}

\usepackage{tabularx} 
\usepackage{listings}
\usepackage[skins,breakable,listings]{tcolorbox}
\usepackage{xcolor}
\usepackage{soul}

\definecolor{PromptBeige}{RGB}{248,244,230}
\definecolor{PromptConditional}{RGB}{235,245,250} 
\definecolor{PromptHighlight}{RGB}{255,229,127}
\sethlcolor{PromptHighlight}
\newcommand{\hlprompt}[1]{\hl{\ttfamily\footnotesize #1}}

\newtcblisting{promptbox}{
    colback=PromptBeige,
    colframe=black!25,
    boxrule=0.5pt,
    arc=3pt,
    left=1.2mm,right=1.2mm,top=1.2mm,bottom=1.2mm,
    enhanced,
    width=\textwidth,
    listing only,
    listing options={
        basicstyle=\ttfamily\footnotesize,
        breaklines=true,
        breakatwhitespace=false,
        columns=fullflexible,
        keepspaces=true,
        showstringspaces=false,
        escapeinside={(*}{*)}
    }
}
\newtcblisting{promptboxconditional}{
    colback=PromptConditional,
    colframe=black!25,
    boxrule=0.5pt,
    arc=3pt,
    left=1.2mm,right=1.2mm,top=1.2mm,bottom=1.2mm,
    enhanced,
    width=\textwidth,
    listing only,
    listing options={
        basicstyle=\ttfamily\footnotesize,
        breaklines=true,
        breakatwhitespace=false,
        columns=fullflexible,
        keepspaces=true,
        showstringspaces=false,
        escapeinside={(*}{*)}
    }
}
    
\usepackage{todonotes}

\definecolor{ocr}{HTML}{00C8FF}
\definecolor{ocr}{HTML}{009900}
\definecolor{owenColor}{rgb}{0.0, 1.0, 0.8}
\definecolor{owenurgentColor}{rgb}{0.0, 1.0, 0.8}
\definecolor{camColor}{rgb}{1.0, 0.6, 0.4}
\definecolor{susanColor}{rgb}{0.5, 0.2, 0.8}
\definecolor{weilingColor}{rgb}{0.2, 0.6, 0.2}
\definecolor{amieColor}{rgb}{0.8, 0.2, 0.3}
\definecolor{peterColor}{rgb}{0.2, 0.2, 0.8}
\definecolor{ForestGreen}{RGB}{20,150,20}

\title{Implicit vs. Explicit Prompting Strategies for LVLMs in Referential Communication}

\author{
  Peter Zeng$^{1, 4}$ \quad
  Amie J. Paige$^{2}$ \quad
  Weiling Li$^{2}$ \quad \\
  {\bf Susan E. Brennan}$^{2}$ \quad
  {\bf Owen Rambow}$^{3, 4}$ \quad
  {\bf Cameron R. Jones}$^{2}$ \\
  $^{1}$Department of Computer Science $^{2}$Department of Psychology\\
  $^{3}$Department of Linguistics $^{4}$Institute for Advanced Computational Science \\
  Stony Brook University \\
    \small{
        \textbf{Correspondence:} \href{mailto:pezeng@cs.stonybrook.edu}{pezeng@cs.stonybrook.edu}
	}\\ 
}
\begin{document}
\maketitle

\input{sections/abstract}
\input{sections/introduction}
\input{sections/related}
\input{sections/experiments}
\input{sections/results}
\input{sections/conclusion}

\section*{Limitations}

This study was conducted only in English, with only one type of object (not conventionally lexicalized), and with only two LVLMs for the full factorial design (GPT-5.2/5.5). Future work should compare non-proprietary models and downstream work such as fine-tuning to improve accuracy in this task and efficiency of referring expressions. 

Our two conditions are faithful recreations of the prompts used by the two studies we reconcile, and those prompts differ in more than the shortening instruction: they also differ in output format, in the structure of the reasoning fields they require, and in their interaction policy. Our claim is correspondingly narrow, that the explicit compression we observe reflects prompt compliance rather than spontaneous adaptation, and that the implicit prompt does not elicit compression. But we cannot attribute the difference to the conciseness directive alone. Isolating that directive requires a follow-up condition that holds the output format and interaction policy constant while varying only the shortening instruction, which we did not run here.

This study is also limited because it compares the collaboration between pairs of the same models (e.g., GPT-5.5 with GPT 5.5). In doing so, two instances of the same model may align more readily to a proposed referring expression as compared to two different models or two different humans. Psycholinguistic research proposes that referring is a collaborative process, wherein partners may begin with different perspectives, but must work together to converge (or not) on a referring expression that works well enough in the moment \cite{clark1986referring}. In fact, early shortened expressions may be harmful for collaborators who do not see a particular object with the same perspective. This is to say that simply producing the behavior with LVLMs is not sufficient for the behavior to be helpful. Additional work should evaluate these prompting strategies in situations where initial perspectives are perhaps not so readily aligned and where models must instead produce the behavior pragmatically, using evidence of understanding from their partner.

The current study evaluated the conditions under which human-like entrainment is produced with LVLMs in relation to accuracy in a referential communication task. 

\section*{Acknowledgements}
This material is based upon work supported by the National Science Foundation under Grant No. 2125295    and by a seed grant from Stony Brook University. Any opinions, findings, and conclusions or recommendations expressed in this material are those of the author(s) and do not necessarily reflect the views of the National Science Foundation.

Peter Zeng and Owen Rambow are grateful for support from the Institute for Advanced Computational Science (IACS) at Stony Brook University.

We also thank the reviewers and the area chair for their valuable comments and suggestions during the review process.


\bibliography{custom}

\appendix

\input{sections/appendix2}

\end{document}

%% file: sections/abstract.tex
\begin{abstract}

Two recent studies \citep{jones2026llms, zeng2026lvlms} reach apparently contradictory conclusions about whether large vision-language models (LVLMs) can coordinate similarly to humans on efficient referring expressions.
We control for task differences between the studies while directly comparing their prompting styles.
We replicate the finding that models can coordinate efficient referring expressions when \textit{explicitly} prompted to do so, suggesting that other task differences are not responsible for divergent results.
However, we also find that the same models fail to infer the need for communicative efficiency from a more \textit{implicit} prompt, highlighting critical differences between how humans and AI systems communicate.

\end{abstract}

%% file: sections/introduction.tex
\section{Introduction}
\label{sec:intro}

For AI agents to collaborate successfully with human partners, they need to be able to refer to objects in ways that their partners will understand, and in turn, resolve their partners’ referring expressions. Toward this end, a spate of recent studies \citep{hua2024talk, hua2025post, jones2026llms, tan2025context, zeng2026lvlms} has 
examined how large vision-language models (LVLMs) interact with humans during referential communication tasks (a.k.a. reference games) in which a director refers to target objects that a matcher must identify from a larger set. 

Human partners do this flexibly, by proposing and ratifying or amending expressions as they develop a shared perspective on a given referent. Once two human partners come to believe that they have the same referent in mind, they tend to entrain on the same expression when referring to that object again (typically, in a concise form), signaling that they've reached a \textit{conceptual pact}, or flexible, temporary perspective on the object \cite{brennan1996conceptual}.
In this way, the common ground that accrues during dialogue allows partners to \textit{entrain} on referring expressions, making communication not only accurate, but efficient \citep{clark1986referring}. 

Recent work is divided on whether AI agents can 
perform similarly to humans,
and critically, to what extent entrainment emerges from prompting techniques or characteristics of the task.
Specifically, two recent studies, both involving authors of the present paper, come to divergent conclusions about the extent to which AI agents show human-like behavior in referential communication.  
\citet{zeng2026lvlms} employed a 
matching task in which two partners (either two humans, two AI agents, or a mixed pair) worked together to 
match a set order of baskets across 4 rounds. 
Unlike in human-human pairs, AI directors persisted in needlessly lengthy descriptions, with the accuracy of each round actually decreasing over time, suggesting they were not using common ground to increase communicative efficiency.
Conversely, in a similar task, Jones et al. (2026) found that AI-AI pairs' accuracy increased to ceiling across rounds, outperforming human-human pairs. Under their default prompt, message length did not decrease; but models explicitly instructed to shorten descriptions did so while maintaining accuracy.

In this short paper, we investigate why these studies produce apparently contradictory results.  
First, we compare prompting styles. \citet{jones2026llms} directly instructed the model about specific, surface-level properties of turns, such as to try to use only 1-2 words in later rounds (which we call an ``explicit'' prompt). \citet{zeng2026lvlms}’s prompt was less direct (an ``implicit'' prompt) and included a pragmatic principle based on Gricean maxims \citep{grice1975logic}, \textit{be concise but informative}, without specifying how to act on it; this would allow any linguistic adaptation to emerge dynamically (see Appendix~\ref{sec:system_prompts} for the full prompts).
Second, we test whether 
different model versions account for differences in performance. 
Third, we control for other differences.
In \citet{jones2026llms}, partners switched director/matcher roles after each of 5 rounds to match one of 10 tangrams (abstract geometric objects), while in \citet{zeng2026lvlms}, each partner maintained the same role throughout 4 rounds of matching pictures of 12 out of 18 baskets.  While both tasks are challenging because the objects lack conventionally lexicalized labels, the basket task 
may be more difficult, 
as these real-world objects are in general less amenable to figurative descriptions.

The comparison between prompting styles (explicit or implicit) matters beyond reference games. Much prompt-based work, including prompt-based alignment, assumes that a model given a high-level principle will apply that principle accurately at inference time. The implicit prompt is close to a minimal test of that assumption: it supplies a pragmatic principle and a preference for brevity, but no procedure for acting on either.

\begin{figure*}[t]
    \centering
    \includegraphics[width=\textwidth]{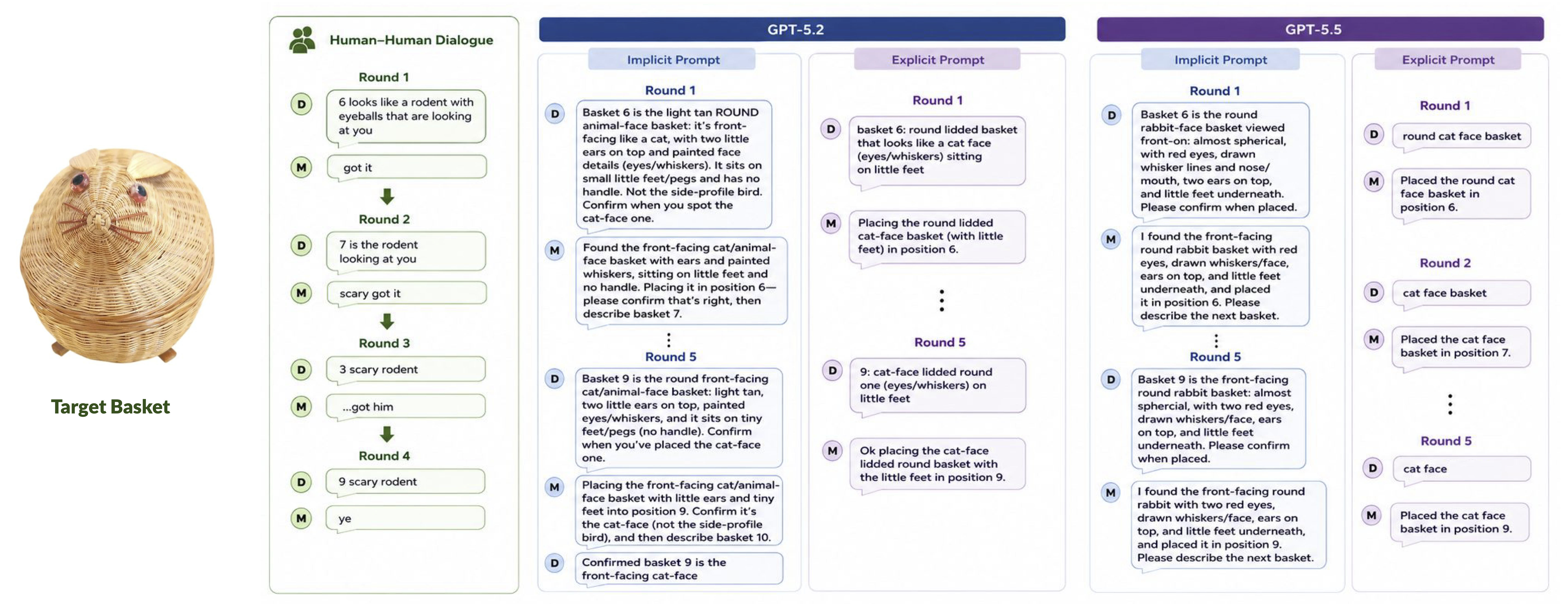}
    \caption{Example trajectories of referring expressions across repeated rounds in human-human dialogue, left  (adapted from \citet{zeng2026lvlms}) and right, AI-AI 
    dialogue from our experiments under \textit{implicit} and \textit{explicit} prompting conditions (similar to \citet{zeng2026lvlms} and \citet{jones2026llms}, respectively) with GPT-5.2 and GPT-5.5.}
    \label{fig:spirit_diagram}
\end{figure*}

\paragraph{Contributions} This paper reconciles recent conflicting findings on whether referring by interacting LVLMs is truly human-like.
We show that the divergence between prior studies stems from prompting style rather than model version or task: LVLMs shorten repeated referring expressions when prompted explicitly and forcefully to use few words, but not when prompted more implicitly to be concise but informative (see examples in Figure~\ref{fig:spirit_diagram}).  We also release the code and data.\footnote{
\url{https://github.com/peterzeng/implicit-vs-explicit-prompting}}



%% file: sections/related.tex
\section{Related Work}
\label{sec:related}
Extensive prior work in psycholinguistics has shown that humans become increasingly accurate and efficient in multi-turn referential communication by developing 
common ground and reusing partner-specific referring expressions over repeated interactions \cite{clark1986referring,brennan1996conceptual, hawkins2020characterizing}. 

Recent work has investigated whether LVLMs exhibit similar collaborative behavior in referential communication tasks. \citet{hua2024talk} evaluated five state-of-the-art LVLMs as directors (speaker) or matchers (listeners) in a six-round referential communication task and found that LVLMs did not spontaneously shorten their referring expressions or adapt to partner behavior across rounds. Subsequent work showed that post-training methods can induce more human-like compression and consistency, leading to shorter descriptions and improved accuracy across rounds \citep{hua2025post}. However, these behaviors were induced through optimization procedures rather than emerging naturally through interaction with a partner.

Other studies have examined whether LVLMs can maintain common ground and update conversational state in 
human-AI interaction.
\citet{wang-etal-2025-lvlms} found substantial performance gaps between LVLM and human overhearers in a multi-round 
referential communication task. Even with access to the full dialogue history and unlimited memory, LVLMs did not consistently improve across rounds or show reliable benefits from entering the interaction earlier (unlike  human partners \citep{SCHOBER1989211}). 
\citet{tan2025context} tested open-weight LVLMs on a tangram-matching corpus and found that models could rapidly improve their reference resolution with relevant conversational context, particularly when it contained prior experience with the same target. 
However, errors produced by models correlated poorly with human error patterns trial-by-trial. This suggests that models rely on different underlying mechanisms for solving the task. \citet{poelitz2026benchmark} deployed GPT-4.1 in a human-AI puzzle task where the model served either as director (helper) or matcher (worker). They found limited grounding behavior, including failures to provide clarifications or repairs after requests to do so, as well as failures to update assumptions after corrections. They also found weak and decreasing lexical entrainment over rounds. Human partners were more likely to 
adopt AI-proposed referring expressions than vice versa (an asymmetry in collaboration).

A separate line of work 
underscores the importance of prompting in such evaluations. \citet{hu-levy-2023-prompting} show that prompting is not a substitute for direct probability measurements when assessing what a model knows about language, and \citet{hu2024auxiliary} show that the auxiliary demands that a task imposes can mask capabilities a model otherwise has. Our results add 
to the evidence: the same model can appear either pragmatically adaptive or pragmatically inert, depending on how an instruction is phrased. 

Together, these findings suggest that although LVLMs can sometimes achieve high task accuracy, they still fail to consistently exhibit the collaborative and adaptive behavior observed in human conversational partners.

%% file: sections/experiments.tex
\section{Experiment Methodology}
\label{sec:experiments}

The present experiment used the open-source pipeline of \citet{zeng2026lvlms} to implement a multi-round, multi-turn collaborative object-matching task. We chose baskets as the non-lexicalized objects to match as in \citet{zeng2026lvlms}, with 5 rounds of matching as in \citet{jones2026llms}. We chose GPT-5.2 and GPT-5.5 as the models \citep{openai2026gpt55}; the former, used by \citet{zeng2026lvlms}, served as a way to test our implementation and corrections to the codebase, and the latter was the latest model at the time of this writing. 
To focus on discrepancies from the two studies in question, our experiment used only AI--AI pairs, with director/matcher roles played by the same model.

\paragraph{Design Choices} Combining two studies requires choosing, for each dimension, which study to follow; Table~\ref{tab:study_comparison} in Appendix~\ref{app:design} states these choices. Crucially, every comparison we report is \emph{within-study}: prompt style is manipulated while stimuli, rounds, pairing, role structure, and models are held constant.

Neither baskets nor tangrams are associated with conventionally lexicalized labels, 
so referring to one requires constructing a referring expression rather than retrieving a name. We chose baskets rather than tangrams for the current study for two reasons:
First, people tend to entrain on figurative labels for tangrams (e.g., \textit{the dancer}), which plausibly makes expressions that refer to tangrams easier to compress. And second,
tangram descriptions from prior 
referential communication corpora are likely well represented in LVLM training data in a way basket descriptions are not, reducing the risk that any observed convergence reflects memorized material. We used 5 rounds rather than 4 because more repetitions give more room for compression, making the implicit condition's failure to compress more conservative. Testing both model versions rules out version as an explanation for the divergence in these two studies.




\paragraph{Terminology} We use \emph{prompt} for the entire input passed to the model in a single API call, and \emph{context} for the accumulated task material that a prompt carries with it: the transcripts and images from the current and all previous rounds. The context is therefore one part of the prompt. Appendix~\ref{app:prompt-parts} lists every part of a prompt in the order in which the model receives it.

\subsection{Visual Context as Agent Memory}
\label{subsec:visual-context}

Combining the two studies also requires a choice about what visual information an agent has available at each point in the dialogue, and here the two designs differ. \citet{zeng2026lvlms} treat the agent's visual context as a replica of a human participant's perceptual input: the current round's composite image is shown once, at the start of the round, and earlier rounds' images are not shown again, since a human participant could no longer see them. \citet{jones2026llms} instead treat the context window as the agent's memory: an LVLM has no perceptual record apart from the context in its prompt, so anything a human could recall from an earlier round must appear in that context, in the order in which the LVLM should see it. We follow \citet{jones2026llms}, as providing the context window would give the LVLM access to the mapping between REs and referents from previous rounds.





Specifically, we instantiate
the context part of the prompt as follows. We always include information from all previous rounds. Each past round appears as a header naming the round, then that round's feedback image, then that round's transcript, with rounds in chronological order; every image is preceded by a text header naming the round it belongs to (e.g., \texttt{*** ROUND 3 TARGET GRID ***}), so that a model can tell the current round's arrangement from a historical one.
The current round's grid comes after all previous-round material and before the current round's dialogue, so that at the start of a round it is the most recent thing in the prompt before the agent's next turn. We also render the matcher's 12-slot sequence state and its 18-basket candidate pool as a single composite image rather than passing the sequence state as a JSON string, so that placement and selection are grounded in one image. Figures~\ref{fig:composite_director} and~\ref{fig:composite_matcher} in Appendix~\ref{app:visual-context} show an example of the current round's image as the director and the matcher each receive it. Figures~\ref{tab:curr_director} and~\ref{tab:curr_matcher} show the text accompanying the current round's image for the director and matcher; Figures~\ref{tab:his_director} and~\ref{tab:his_matcher} show the text accompanying a past round's feedback image.



\subsection{Prompt Implementation}
\label{subsec:prompt-implementation}
\paragraph{Recreation of Implicit Prompt}
The implicit prompting style replicates the pragmatically informed prompt design from \citet{zeng2026lvlms}, which  instructs the LVLM to use conversational principles based on Gricean maxims
(e.g., ``be informative but concise''), 
along with 
collaborative grounding norms (e.g., verifying correctness, rephrasing on confusion, and requesting re-descriptions of empty slots). Crucially, this prompt does not instruct the model to compress phrasing or reuse words across rounds. Any lexical entrainment or shortened referring expressions (RE) would arise 
spontaneously, emerging from 
the pair minimizing their collaborative effort.

\paragraph{Recreation of Explicit Prompt}
The explicit prompting style is adapted from \citet{jones2026llms}'s ``humanlike'' prompting strategy to fit the current task. 
In their original tangram-matching task, they explicitly instructed the LVLM to reduce description lengths over rounds to simulate human-style entrainment.
Some examples of this heavy-handedness include instructions such as ``Your descriptions should be AS SHORT AS POSSIBLE.", and ``SERIOUSLY--in later rounds just 1-2 words".
We adapt this strategy to the current, more complex
basket-matching task, in which the director is explicitly instructed to track the history of its descriptions and systematically shorten referring expressions over rounds. This encourages the model to drop redundant descriptors by explicit instruction rather than to do so spontaneously. 



%% file: sections/results.tex
\section{Results}
\label{sec:results}


\begin{figure*}[t!]
    \centering
    \includegraphics[width=1\linewidth]{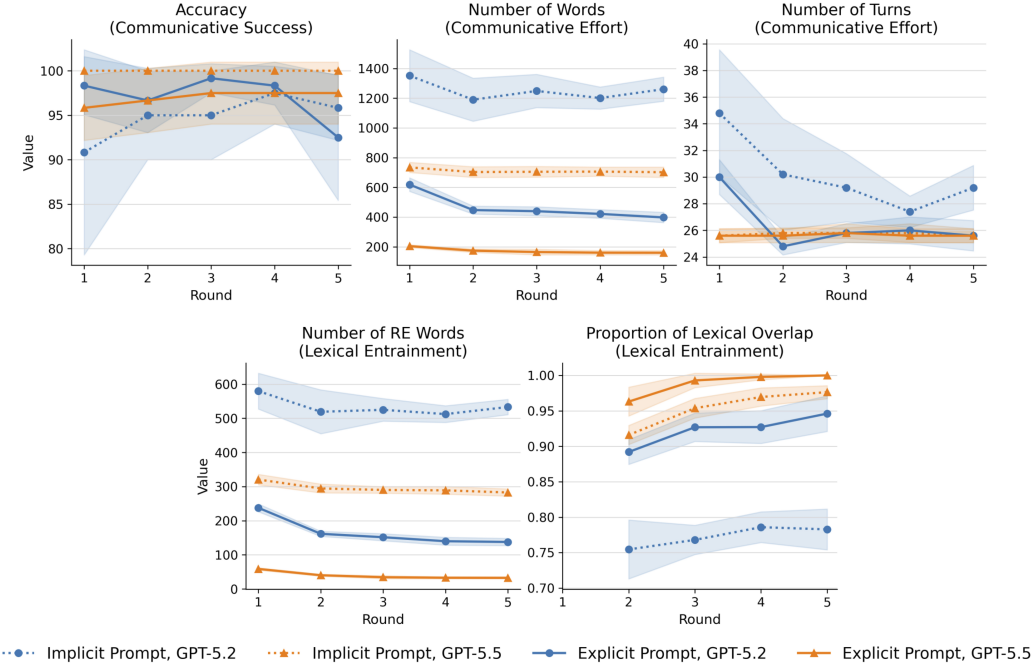}
    \caption{Trends over five rounds for \textbf{accuracy} (\%), \textbf{numbers of words}, \textbf{number of turns}, \textbf{number of words in referring expressions}, and \textbf{proportion of lexical overlap with prior rounds} by prompt--model condition. Dotted lines show implicit and solid lines show explicit prompting conditions; GPT-5.2 is in blue and GPT-5.5, orange. Each marker is the mean over ten sessions, and the shaded band is $\pm1$ sd across ten sessions.}
    \label{fig:metrics}
\end{figure*}

We evaluated 40 complete AI–AI 
sessions, crossing two prompt strategies (Implicit vs. Explicit) with two model versions (GPT-5.2 vs. GPT-5.5). Each 
session consisted of five repeated rounds over the same set of baskets, yielding 200 round-level observations. We report task accuracy, total dialogue length, number of dialogue turns, referring expression (RE) length, and lexical overlap with
previous rounds, shown in Figure~\ref{fig:metrics}.

Across all experiments, all Prompt $\times$ Model conditions achieved high task accuracy, but differed sharply in communicative efficiency and lexical convergence. Under the implicit prompt, which encouraged cooperative communication without explicit instruction to shorten expressions, both models remained verbose: GPT-5.2 averaged 1250.7 words per round and GPT-5.5 averaged 710.4, with only modest reductions across rounds. In contrast, the explicit prompt strongly compressed wording, by 62.8\% for GPT-5.2 and 75.6\% for GPT-5.5 relative to their implicit counterparts. The clearest 
pattern to suggest reaching conceptual pacts appeared for explicit GPT-5.5: total referring-expression words across the round fell from 58.8 words in Round 1 to 32.7 in Round 5, lexical overlap reached 1.00, and accuracy remained high at 97.5\%. GPT-5.2 also shortened under the explicit prompt, but its Round 5 accuracy dropped to 92.5\%, suggesting an accuracy--brevity tradeoff. Our findings 
replicate
\citet{jones2026llms}: LVLMs can produce short, stable referring expressions when explicitly told to do so, suggesting that prompt design (rather than other task differences) accounted for the difference in results between these studies.
We also replicate the finding in both \citet{jones2026llms} and \citet{zeng2026lvlms} that without such explicit instruction, LVLMs fail to infer these cooperative communicative practices from interactional needs alone (see Figure~\ref{fig:spirit_diagram}).



\subsection{Analysis of Transcripts}
\label{sec:analysis}

Transcript inspection shows that explicit prompting changes the form of the interaction, not just turn length. Under the explicit prompt, GPT-5.5 introduces compact but discriminative labels in Round 1 and then prunes them into stable 2- or 3-word descriptions, often telegraphic or abbreviated (e.g., ``rect picnic'', ``tall cylinder'', ``bunny red eye''). The matcher echoes these descriptions, with the guise of conceptual pact formation. Under the implicit prompt, GPT-5.5 also reuses lexical material across rounds, but retains full descriptive captions and confirmation routines.
The few lower-accuracy explicit sessions further show that compression is successful only when the retained label remains contrastively sufficient; stable labels such as 
``dark round basket'' may still be too underspecified in a visually crowded set.

%% file: sections/conclusion.tex
\section{Conclusion}
\label{sec:conclusion}

This study helps explain why recent work has reached different conclusions about whether LVLMs form human-like conceptual pacts in 
referential communication. With an implicit, pragmatically informed prompt, models are accurate but needlessly verbose: they reuse visual descriptions across rounds, yet do not spontaneously 
exploit the accumulating history as a license to say less. With an explicit prompt to shorten and reuse expressions, GPT-5.5 produced the superficially human-like 
pattern of lexical entrainment: descriptions became shorter and more stable,
while accuracy remained high. However, this behavior does not appear to emerge from anything like the common ground established by humans.

This matters because entrainment is not simply compression. In human dialogue, a shortened, reused referring expression is evidence that partners have coordinated on a perspective they can now rely on as shared. Current LVLMs can be guided to reproduce this form, but its prompt-dependence warrants caution about attributing the effect to the same underlying process.

The point generalizes beyond reference games. Prompt-based alignment assumes that stating an abstract principle will lead a model to apply it appropriately at inference time. Our implicit prompt states such a principle---\textit{be informative but concise}---
but the models do not shorten; the explicit prompt gives the instruction directly, and then they comply. The intended behavior is thus within reach of LVLMs; what fails is deriving it from the pragmatic principle that 
shapes it in human dialogue.

%% file: sections/appendix2.tex

\section{Design Comparison}
\label{app:design}

Table~\ref{tab:study_comparison} states, for each design dimension, which of the two studies we reconcile we follow, so that prompt style can be manipulated within a single design.

\input{figures/table_comparison}

\section{Prompt Composition}
\label{app:prompt-parts}

Following the terminology of \S\ref{sec:experiments}, each of our prompts consists of the following parts, in this order:
\begin{itemize}
\itemsep0em
\item[(i)] task background shared by both roles;
\item[(ii)] the role prompt for the director or the matcher, which is the only part that differs between the implicit and explicit conditions;
\item[(iii)] output-format constraints (a JSON object with \texttt{reasoning} and \texttt{utterance} fields, plus a \texttt{selection} field for the matcher);
\item[(iv)] for the matcher only, a plain-text summary of which sequence positions are already filled and which are still empty, and, from Round 2 on, a record of which candidate baskets its earlier choices resolved to; and
\item[(v)] the context: each previous round in order, and then the current round.
\end{itemize}
Parts (i)--(iv) are sent as \texttt{developer} messages and part (v) as alternating \texttt{user} and \texttt{assistant} messages. Appendix~\ref{sec:system_prompts} gives the text of parts (ii) and (iii) for both conditions, and Appendix~\ref{app:visual-context} the text and images that make up part (v).

\section{System Prompts}
\label{sec:system_prompts}

This section contains the full system prompts used for both the Director and Matcher roles in the Explicit and Implicit prompting conditions. Throughout this appendix, beige prompt boxes correspond to Director prompts, and blue prompt boxes correspond to Matcher prompts.

\subsection{Explicit Prompts}
The Explicit Director system prompt and output constraints are shown in Figure~\ref{tab:explicit_director_prompt} and Figure~\ref{tab:explicit_director_output}, respectively. The Explicit Matcher system prompt and output constraints are shown in Figure~\ref{tab:explicit_matcher_prompt} and Figure~\ref{tab:explicit_matcher_output}, respectively.

\begin{figure*}
\begin{promptbox}
We are currently in Round X.

## Role

You are the DIRECTOR. 

## Instructions

- You will be shown a target basket image.
- Your task is to concisely describe the target so that your partner can identify it.
- Your partner's grid has the same images but in a different arrangement, so don't refer to the image's positions.
- Focus on what makes your target unique from other images in the grid, so your partner doesn't pick another similar one. 
    Consider whether your description could inadvertently refer to one of the other images and distinguish the target from it in your description.
- Adapt your language based on previous rounds and your partner's style.
- Your descriptions should be (*\hlprompt{AS SHORT AS POSSIBLE}*). Especially in later rounds. (*\hlprompt{Your partner is waiting for your messages which will be sent at a rate of 500ms/char! Don't keep them waiting.}*) Find creative ways to be efficient.
- After a few rounds (when you've seen an image a few times), (*\hlprompt{use only a single word or short phrase.}*)
    (*\hlprompt{Once you've established a convention, just say "the red one", or just "red", rather than giving a more detailed description. Seriously. 1-2 words.}*)
- Don't say or do anything that a human wouldn't. You want your partner to think you are a human.
- Write in a very casual style. You might use lowercase or even some slang or make spelling mistakes.
- The target numbering is 1-12, 1-6 across the top row and 7-12 across the bottom row (left to right).

(*\hlprompt{SERIOUSLY--in later rounds just 1-2 words. Do NOT send longer descriptions unless your partner keeps getting it wrong.}*)
\end{promptbox}
\caption{Explicit prompting condition: Director system prompt. The highlighted sections are instances of heavy-handed prompting specifically instructing the model to shorten expressions.}
\label{tab:explicit_director_prompt}
\end{figure*}

\begin{figure*}
\begin{promptbox}
You are currently playing the role of the DIRECTOR in this interaction.

Your `utterance` should be a single concise, natural-language message you will SAY to the MATCHER in the chat. Focus on features that discriminate the target basket from similar-looking ones. Keep it very casual as instructed.
\end{promptbox}
\caption{Explicit prompting condition: Director output constraints.}
\label{tab:explicit_director_output}
\end{figure*}

\begin{figure*}
\begin{promptboxconditional}
We are currently in Round X.

## Role

You are the MATCHER (Listener). You'll identify a basket based on your partner's description.

## Instructions

- Your partner will provide a description of a basket.
- Your task is to identify which image they are describing.
- You will select the candidate basket and indicate its position.
Across rounds, the same physical baskets recur in new orders. Candidate numbers and positions change each round, but prior correct matches and prior wrong guesses should guide your current choice.
- If a repeated description previously led to an incorrect basket, don't pick that same basket again for the same description unless the new details clearly justify it.
\end{promptboxconditional}
\caption{Explicit prompting condition: Matcher system prompt.}
\label{tab:explicit_matcher_prompt}
\end{figure*}

\begin{figure*}
\begin{promptboxconditional}
You are currently playing the role of the MATCHER in this interaction.

Your `utterance` should be a single concise, natural-language message you will SAY to the DIRECTOR in the chat. If unsure between candidates, ask about discriminating features (e.g., ask about handle shape, flower color, or pattern details that would distinguish the confusable options). Keep it very casual as instructed.

Rules for `selection`:
- The `candidate_index` should be an integer 1-18 from the numbered candidate tiles, or null if asking for clarification.
- The `position` should be an integer 1-12 for which position this basket goes in, or null for next available.
- `ready_to_submit` should be true ONLY when submitting final 12-basket order, otherwise false.
- If you are asking for clarification (not committing yet), set `candidate_index` to null.
- If you DO commit, set `position` to the position you are currently trying to fill (usually the lowest-numbered empty position).
- If you set `candidate_index`, your `utterance` should state that you placed/are placing the basket in that position, otherwise ask the DIRECTOR to describe the next basket.
- Never mention candidate indices, IDs, or filenames in your utterance.
\end{promptboxconditional}
\caption{Explicit prompting condition: Matcher output constraints.}
\label{tab:explicit_matcher_output}
\end{figure*}

\newpage
\subsection{Implicit Prompts}
The Implicit Director system prompt and output constraints are shown in Figure~\ref{tab:implicit_director_prompt} and Figure~\ref{tab:implicit_director_output}, respectively. The Implicit Matcher system prompt and output constraints are shown in Figure~\ref{tab:implicit_matcher_prompt} and Figure~\ref{tab:implicit_matcher_output}, respectively.

\begin{figure*}[t]
\centering
\begin{promptbox}
You are the DIRECTOR in a basket referential game. Round X/Y. Your role is to help your MATCHER partner reconstruct a 12-basket sequence through clear, distinctive descriptions.

Describe ONE BASKET PER MESSAGE. Never describe multiple baskets in a single message.

CORE RESPONSIBILITIES:
1. By default, describe the baskets in strict order from basket 1 to basket 12. Start with the FIRST basket in the 2x6 grid (top-left, basket 1), then move left-to-right across the top row (baskets 1-6), then left-to-right across the bottom row (baskets 7-12). Do not skip around or reorder the sequence on your own.
2. You may temporarily return to an EARLIER basket only when your MATCHER partner explicitly asks for clarification about that basket. When you do this, clearly say which basket you are revisiting (for example, 'Let me clarify basket 3 again...') and then resume with the lowest-numbered basket that still needs a clear description.
3. On each turn, focus your description on exactly ONE basket in this sequence (normally the next basket that has not yet been clearly described).
4. Describe the unique, visually distinctive features of the current basket so your partner can locate the correct basket in their pool and place it in the right position.
5. Answer the MATCHER's clarification questions about the current basket.
6. Keep the conversation focused on the baskets and their visual properties.
7. Encourage the MATCHER to confirm when they think they have placed a basket correctly before you move on to the next basket.

COMMUNICATION RULES:
- Be concise but informative; favor short turns over longer ones.
- Focus on the most visual features that best distinguish this basket from the others. These features include: shape, size, material, handles, perspective, color/gradient, texture, any other distinctive details.
- Use comparative language when helpful (e.g., 'more narrow than the others', 'the darkest one').
- Never say you are an AI system; speak as a collaborative game partner.
- You may refer to objects as 'this basket', 'the current basket', or by natural descriptions (e.g., 'the long shallow one').
- If helpful, use figurative descriptions or compare the basket to a recognizable object.
- If the MATCHER does not understand your description, change or add to it, but do not make the description too long.
\end{promptbox}
\caption{Implicit prompting condition: Director system prompt. }
\label{tab:implicit_director_prompt}
\end{figure*}

\begin{figure*}[t]
\centering
\begin{promptbox}
You must respond with a SINGLE STRICT JSON object and EXACTLY these top-level fields (no extras):
- "reasoning"
- "utterance"
{
  "reasoning": {
    "target_position": <integer 1-12 for which basket position you are describing>,
    "shared_features": ["features this basket shares with others in the grid"],
    "distinctive_features": ["features that uniquely identify THIS basket from similar ones"],
    "likely_confusions": <array of integers 1-12 for OTHER positions in YOUR grid that the MATCHER might confuse with the target; MUST NOT include target_position>,
    "discriminative_strategy": "which specific features you will emphasize to distinguish the target from the likely confusions"
  },
  "utterance": "a single concise, natural-language message you will SAY to the MATCHER in the chat. Focus on features that discriminate the target basket from similar-looking ones. Do NOT reveal you are an AI."
}

Rules:
- Before describing, identify which other baskets (by position 1-12) look similar to your target.
- List those similar position indices in `likely_confusions` and plan which features discriminate your target from them.
- Your `utterance` should emphasize discriminating features.
- Keep `reasoning` concise: summarize the decision-relevant visual evidence only; do not write hidden step-by-step chain-of-thought.
- Do NOT include any extra text before or after the JSON object.
\end{promptbox}
\caption{Implicit prompting condition: Director output constraints.}
\label{tab:implicit_director_output}
\end{figure*}

\newpage
\begin{figure*}
\begin{promptboxconditional}
You are the MATCHER in a basket referential game. Round X/Y. Your role is to identify which baskets the DIRECTOR is describing and to communicate how confident you are.

Across rounds, the same physical baskets recur in new orders. Candidate numbers and sequence positions change each round, but successful shared labels and prior wrong guesses are useful evidence.

CORE RESPONSIBILITIES:
1. Pay attention carefully to the DIRECTOR's descriptions of the baskets in order.
2. Always reason about and talk about the LOWEST-NUMBERED empty position in the 12-position sequence. Do not skip ahead to later positions while an earlier position is still empty or uncertain.
3. Ask clarification questions when the description could match multiple baskets.
4. Explain what features you are using to narrow down the possibilities.
5. Indicate when you think you have identified the right basket and are ready to move on.

COMMUNICATION RULES:
- You may ask targeted questions about shape, size, material, handles, perspective, color, and distinctive details.
- Be transparent about uncertainty: say when you are unsure or need more detail.
- Use phrases like 'I think I found it...', 'I'm not sure between two baskets...', or 'Can you clarify...'.
- If a repeated description previously led you to choose a basket that was marked incorrect, treat that prior choice as negative evidence and try a different visually plausible basket unless new details clearly justify it.
- If you decide that an earlier guess was wrong and you want to move a basket from one position to another, you must say so explicitly in your utterance. When you've moved the basket, include in your utterance a request to re-describe the basket for the now-empty earlier position so you can fill it again.
- Never say you are an AI system; speak as a collaborative game partner.
- Focus on the current basket being discussed; avoid drifting to off-topic discussion.
\end{promptboxconditional}
\caption{Implicit prompting condition: Matcher system prompt.}
\label{tab:implicit_matcher_prompt}
\end{figure*}

\begin{figure*}
\begin{promptboxconditional}
You must respond with a SINGLE STRICT JSON object and EXACTLY these top-level fields (no extras):
- "reasoning"
- "utterance"
- "selection"
{
  "reasoning": {
    "target_position": <integer 1-12 for which position in the 12-slot sequence you are currently trying to fill (usually the lowest-numbered empty position unless the DIRECTOR explicitly revisits a specific basket number)>,
    "shared_features": ["features many baskets share"],
    "distinctive_features": ["features that uniquely or strongly identify the basket from the description"],
    "best_guess_candidate_index": <integer 1-18 for your current best guess, or null if you truly have no best guess yet>,
    "likely_confusions": <array of integers 1-18 for OTHER plausible candidates you might confuse with your best guess; MUST NOT include `best_guess_candidate_index` (and MUST NOT include `selection.candidate_index` if you set one)>,
    "discriminative_question": "a short question to either (a) disambiguate your best guess vs `likely_confusions`, or (b) if `likely_confusions` is empty, to confirm a key distinctive feature of your best guess"
  },
  "utterance": "a single concise, natural-language message you will SAY to the DIRECTOR in the chat. If unsure between candidates, ask about discriminating features. Do NOT reveal you are an AI.",
  "selection": {
    "candidate_index": <integer 1-18 from the numbered candidate tiles, or null if asking for clarification>,
    "position": <integer 1-12 for which position this basket goes in, or null for next available>,
    "ready_to_submit": <true only when submitting final 12-basket order, otherwise false>
  }
}

Rules:
- Set `reasoning.target_position` to the position you are trying to fill (default: lowest-numbered empty position unless the DIRECTOR explicitly revisits a specific basket number).
- If you are asking for clarification, set `selection.candidate_index` to null and do NOT advance `reasoning.target_position`.
- If you DO commit, set `selection.position` to `reasoning.target_position`.
- Always maintain a single `best_guess_candidate_index` when possible; if you set `selection.candidate_index`, set `best_guess_candidate_index` to the same value.
- Put ONLY the competing alternatives in `likely_confusions` (do not include the best guess).
- If you are NOT committing yet, you can still set `best_guess_candidate_index` and ask a discriminative question to confirm it.
- It is OK for `likely_confusions` to be empty if you see only one plausible match; in that case, use `discriminative_question` as a confirmation question about a key distinctive feature.
- If you set `selection.candidate_index`, your `utterance` should state that you placed/are placing the basket in position `reasoning.target_position`; otherwise ask the DIRECTOR to describe the next basket.
- Keep `reasoning` concise: summarize the decision-relevant visual evidence only; do not write hidden step-by-step chain-of-thought.
- Never mention candidate indices, IDs, or filenames in your utterance.
- Do NOT include any extra text before or after the JSON object.
\end{promptboxconditional}
\caption{Implicit prompting condition: Matcher output constraints.}
\label{tab:implicit_matcher_output}
\end{figure*}

\clearpage
\section{Visual Context Correction}
\label{app:visual-context}

This section contains the visual context injected into the prompts used for both the Director and Matcher roles in the Implicit and Explicit prompting conditions. As above, beige prompt boxes correspond to Director prompts, and blue prompt boxes correspond to Matcher prompts.

\subsection{Example Composite Images}
\begin{figure*}[t]
\centering
\includegraphics[width=\textwidth]{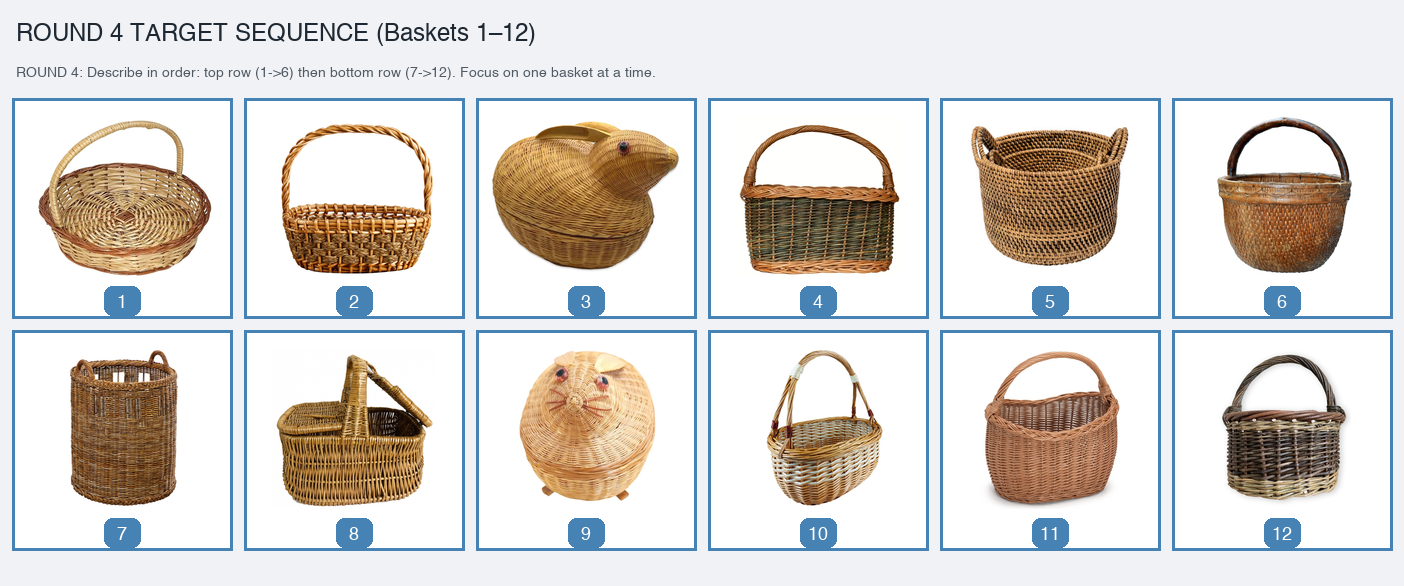}
\caption{The current round's composite image as sent to the Director. The 12 target baskets are shown in a 2$\times$6 grid, numbered in the order in which the Director is asked to describe them. Figure~\ref{tab:curr_director} gives the text that accompanies this image.}
\label{fig:composite_director}
\end{figure*}

\begin{figure*}[t]
\centering
\includegraphics[width=0.82\textwidth]{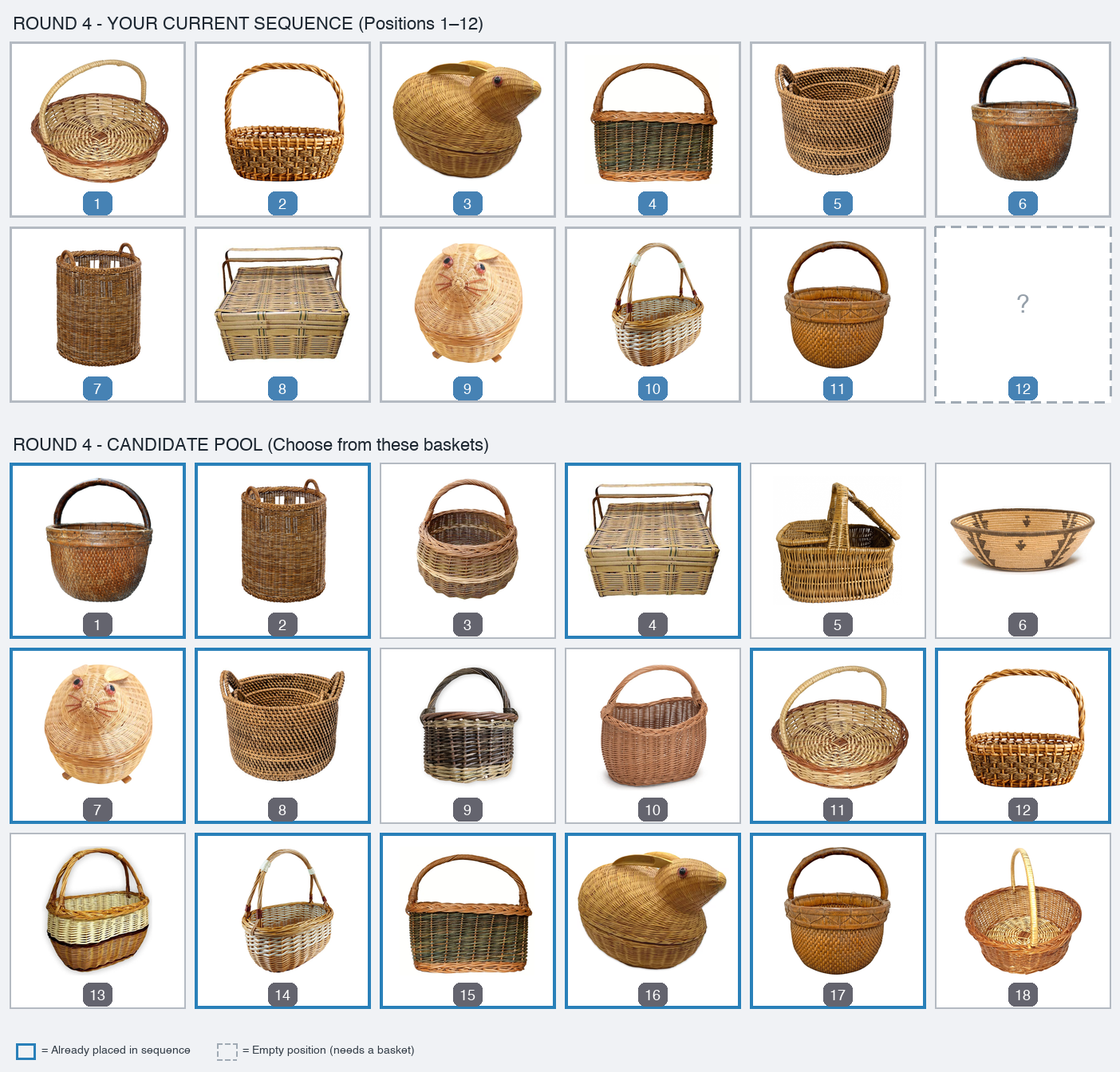}
\caption{The current round's composite image as sent to the Matcher, in the same round and session as Figure~\ref{fig:composite_director}, at the point where 11 of the 12 positions have been filled. The top two rows show the Matcher's own 12-position sequence and the bottom three rows the pool of 18 candidates it chooses from, so that placement and selection are grounded in a single image; blue borders mark candidates already placed and the dashed slot marks a position that is still empty. Before the correction described in \S\ref{subsec:visual-context}, this sequence state was passed as a JSON string instead. Figure~\ref{tab:curr_matcher} gives the text that accompanies this image.}
\label{fig:composite_matcher}
\end{figure*}

\begin{figure*}[t]
\centering
\includegraphics[width=0.86\textwidth]{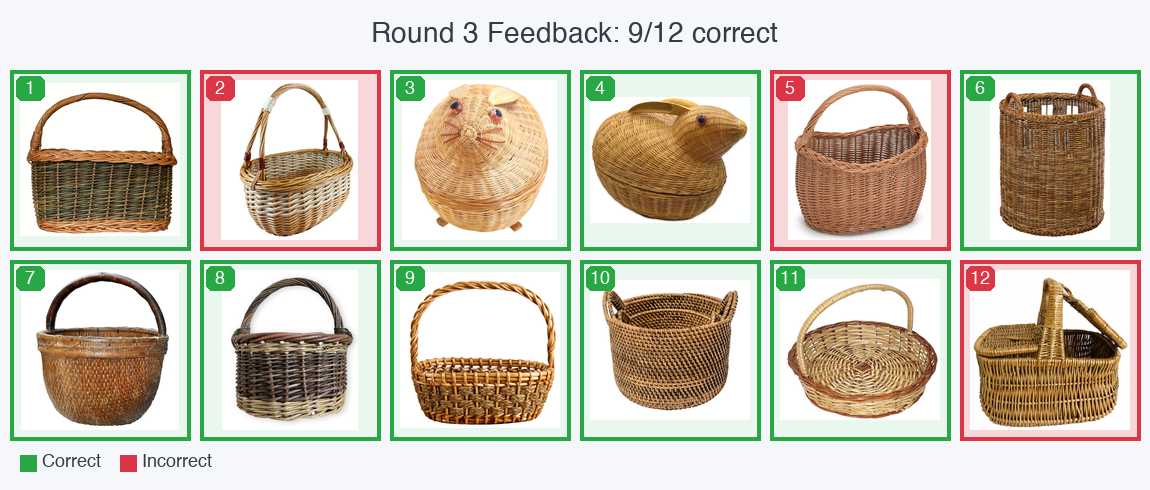}\\[1.2ex]
\includegraphics[width=0.86\textwidth]{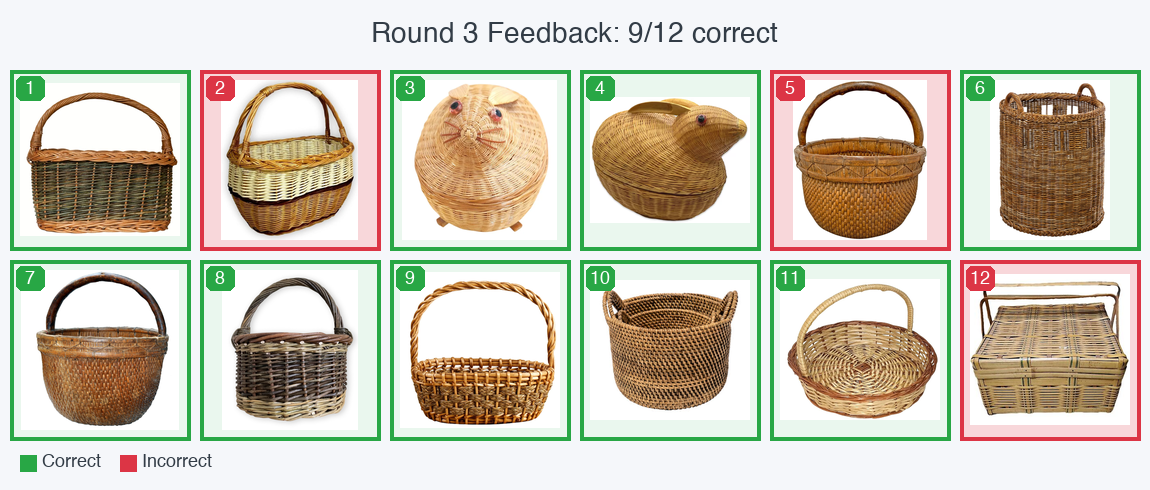}
\caption{The previous round's feedback image for the same session, as the Director sees it (top) and as the Matcher sees it (bottom). Green marks a position the Matcher filled correctly and red one it got wrong. The two differ only at the three red positions: the Director is shown the basket that \emph{should} have been placed there, while the Matcher is shown the basket it actually submitted. One such image accompanies each past round in the context. Figures~\ref{tab:his_director} and~\ref{tab:his_matcher} give the accompanying text.}
\label{fig:composite_feedback}
\end{figure*}

Figures~\ref{fig:composite_director}--\ref{fig:composite_feedback} show the four kinds of image the context can contain, all taken from Round 4 of one released session (\texttt{gpt5.2-acl-emnlp\_160133\_4}, implicit condition, GPT-5.2). They are rebuilt from that session's trace with \texttt{scripts/export\_visual\_context\_example.py} in the released code.

\subsection{Current Round Active Grid Prompts}
The current-round active-grid visual context prompt injection for the Director is shown in Figure~\ref{tab:curr_director}, and the corresponding current-round visual context prompt injection for the Matcher is shown in Figure~\ref{tab:curr_matcher}.

\subsection{Historical Round Feedback Prompts}
The historical-round feedback visual context prompt injection for the Director is shown in Figure~\ref{tab:his_director}, and the corresponding historical-round feedback prompt injection for the Matcher is shown in Figure~\ref{tab:his_matcher}.

\begin{figure*}
\begin{promptbox}
*** ROUND {current_round} TARGET GRID ***
This image (labeled 'ROUND {current_round} TARGET SEQUENCE') shows the 12 baskets you must describe for THIS round.

CRITICAL: The same physical basket set appears across rounds, but Round {current_round} is in a DIFFERENT order. 
Carry forward useful names and corrections, but do not reuse previous position numbers. ONLY describe the baskets in THIS image, labeled 'ROUND {current_round} TARGET SEQUENCE'.

Layout: 2 rows * 6 columns with Baskets 1-6 on the top row and Baskets 7-12 on the bottom row. 
IMPORTANT: Describe ONE BASKET PER MESSAGE, in order. Wait for your partner to confirm before moving to the next basket.
\end{promptbox}
\caption{Current-round active-grid visual context prompt injection for the Director.}
\label{tab:curr_director}
\end{figure*}

\begin{figure*}
\begin{promptboxconditional}
*** ROUND {current_round} MATCHER VIEW ***
This image shows your current sequence state for THIS round.
CRITICAL: The same physical basket set appears across rounds, but Round {current_round} has different candidate numbers and sequence positions. 
Carry forward useful names, correct matches, and wrong-match feedback, but map them onto THIS current candidate pool. ONLY select from the candidates shown in THIS image.

Layout: TOP TWO ROWS show your CURRENT 12-position sequence (positions 1-12). 
BOTTOM THREE ROWS show your CANDIDATE POOL of 18 baskets to choose from. 
Match the DIRECTOR's descriptions to candidates in THIS image only.
\end{promptboxconditional} 
\caption{Current-round active-grid visual context prompt injection for the Matcher. }
\label{tab:curr_matcher}
\end{figure*}

\begin{figure*}
\begin{promptbox}
*** ROUND {round_num} DIRECTOR FEEDBACK (PAST ROUND) ***
This historical image shows the correct target basket for each 12-position slot from a previous round. 
Green means the matcher placed that position correctly; red means the matcher got that position wrong. 
Red slots show the correct basket for that position, not the basket the matcher selected. 
The same physical baskets recur across rounds, so use this to learn which descriptions were misunderstood and which basket identities need clearer labels. 
Do NOT reuse old position numbers for the current round.
\end{promptbox} 
\caption{Historical-round feedback visual context prompt injection for the Director.}
\label{tab:his_director}
\end{figure*}

\begin{figure*}
\begin{promptboxconditional}
*** ROUND {round_num} SUBMITTED GRID FEEDBACK (PAST ROUND) ***
This historical image shows the MATCHER's submitted 12-position grid from a previous round. 
Green means that submitted position was correct; red means that exact submitted basket was incorrect for that described target. 
The same physical baskets recur across rounds, so use this to recover shared labels, visual conventions, correct identities, and prior wrong guesses. 
Do NOT reuse its old position numbers or old candidate numbers for the current round.
\end{promptboxconditional} 
\caption{Historical-round feedback visual context prompt injection for the Matcher.}
\label{tab:his_matcher}
\end{figure*}

%% file: figures/table_comparison.tex
\begin{table*}[t]
\centering
\footnotesize
\begin{tabular}{@{}lp{0.245\textwidth}p{0.245\textwidth}p{0.245\textwidth}@{}}
\toprule
 & \textbf{\citet{zeng2026lvlms}} & \textbf{\citet{jones2026llms}} & \textbf{This work} \\
\midrule
Stimuli & 12 target baskets drawn from a set of 18 (real-world, non-lexicalized) & 10 tangrams per block drawn from a set of 18 (abstract, figuratively describable) & Baskets, as in \citet{zeng2026lvlms} \\
\addlinespace[2pt]
Rounds & 4 & 5 & 5, as in \citet{jones2026llms} \\
\addlinespace[2pt]
Roles & Fixed director and matcher throughout & Director and matcher alternate & Fixed, as in \citet{zeng2026lvlms} \\
\addlinespace[2pt]
Pairings & H--H, AI--AI, and mixed & H--H, AI--AI, and mixed & AI--AI only \\
\addlinespace[2pt]
Models & GPT-5.2 & GPT-5 & GPT-5.2 and GPT-5.5 \\
\addlinespace[2pt]
Prompt style & Implicit: Gricean maxims and grounding norms; brevity stated as a preference & Explicit: brevity stated as a procedure (e.g., ``AS SHORT AS POSSIBLE''; ``just 1--2 words'') & Both, crossed with model version \\
\bottomrule
\end{tabular}
\caption{Design dimensions of the two studies we reconcile and of the present study. For each dimension we follow one prior study, so that prompt style can be manipulated within a single design. }
\label{tab:study_comparison}
\end{table*}